\documentclass{article}

\usepackage[preprint]{neurips_2024}

\usepackage[utf8]{inputenc} 
\usepackage[T1]{fontenc}    
\usepackage{hyperref}      
\usepackage{url}           
\usepackage{booktabs}     
\usepackage{amsfonts}      
\usepackage{nicefrac}      
\usepackage{microtype}     

\PassOptionsToPackage{table,dvipsnames}{xcolor}

\usepackage{microtype}
\usepackage{graphicx}
\usepackage{subcaption}
\usepackage{booktabs} 
\usepackage{arydshln} 
\usepackage{tikz}
\usetikzlibrary{bayesnet}
\usetikzlibrary{arrows}
\usepackage{amsmath}
\usepackage{amssymb}
\usepackage{mathtools}
\usepackage{amsthm}
\usepackage{fontawesome5}
\usepackage{makecell}

\usepackage{float}

\newfloat{floatalgorithm}{htbp}{lop}
\floatname{floatalgorithm}{Algorithm}

\usetikzlibrary{decorations.pathreplacing}

\usepackage{enumitem,multirow,adjustbox}

\usepackage{hyperref}
\usepackage{url}

\usepackage[table]{xcolor}		

\definecolor{kleinblue}{rgb}{0,0.18,0.65}
\definecolor{cite_color}{RGB}{190,0,60}
\definecolor{link_color}{RGB}{153, 0, 0}
\definecolor{url_color}{RGB}{105, 33, 158}
\definecolor{deeppurple}{rgb}{0.5, 0, 0.5} 
\definecolor{burgundy}{RGB}{128, 0, 32}

\hypersetup{colorlinks=true,citecolor=kleinblue, linkcolor=kleinblue, urlcolor=kleinblue}

\definecolor{note}{RGB}{153, 0,0}  
\definecolor{module}{rgb}{0, 0, 0.5}
\definecolor{args}{RGB}{0, 128, 0}
\newcommand{\note}[1]{\textcolor{note}{#1}}

\definecolor{deepgreen}{RGB}{0,100,0}
\definecolor{orange}{RGB}{255,165,0}

\definecolor{brightred}{RGB}{255,51,51}

\newcommand{\best}[1]{\textbf{\textcolor{brightred}{#1}}}
\newcommand{\second}[1]{\textbf{\textcolor{deeppurple}{#1}}}
\newcommand{\third}[1]{\textbf{\textcolor{orange}{#1}}}
\usepackage{bbding}

\definecolor{lightred}{RGB}{255, 200, 200}

\usepackage{soul}

\usepackage{wrapfig}

\usepackage{pifont}

\usepackage{enumitem}
\usepackage{algorithm}
\usepackage{algcompatible}

\usepackage{fontawesome5}

\usepackage{xspace}

\makeatletter
\newcommand*\rel@kern[1]{\kern#1\dimexpr\macc@kerna}
\newcommand*\widebar[1]{
  \begingroup
  \def\mathaccent##1##2{
    \rel@kern{0.8}
    \overline{\rel@kern{-0.8}\macc@nucleus\rel@kern{0.2}}%
    \rel@kern{-0.2}
  }
  \macc@depth\@ne
  \let\math@bgroup\@empty \let\math@egroup\macc@set@skewchar
  \mathsurround\z@ \frozen@everymath{\mathgroup\macc@group\relax}
  \macc@set@skewchar\relax
  \let\mathaccentV\macc@nested@a
  \macc@nested@a\relax111{#1}
  \endgroup
}
\makeatother

\usepackage[hyperpageref]{backref}

\renewcommand*{\backrefalt}[4]{
  \ifcase #1 \relax
  \or
    (Cited on page #2)
  \else
    (Cited on pages #2)
  \fi
}

\definecolor{Gray}{gray}{0.9}

\graphicspath{{./fig/}}

\newcommand{\ourmethod}{CDT}
\newcommand{\ie}{\emph{i.e.}}

\usepackage{lipsum}

\usepackage{caption}

\newcommand{\eg}{\emph{e.g.}}

\title{Rethinking Video Tokenization:\\A Conditioned Diffusion-based Approach}

\author{
Nianzu Yang\textsuperscript{1,{$\dagger$},$*$},
Pandeng Li\textsuperscript{2,$\dagger$},
Liming Zhao\textsuperscript{2},
Yang Li\textsuperscript{1},
Chen-Wei Xie\textsuperscript{2},
Yehui Tang\textsuperscript{1},\\
\textbf{Xudong Lu}\textsuperscript{1},
\textbf{Zhihang Liu}\textsuperscript{2},
\textbf{Yun Zheng}\textsuperscript{2},
\textbf{Yu Liu}\textsuperscript{2},
\textbf{Junchi Yan}\textsuperscript{1,$\S$}
\\[0.4ex]
\textsuperscript{1~}School of Artificial Intelligence \& School of Computer Science, Shanghai Jiao Tong University\quad \\
\textsuperscript{2~}Tongyi Lab, Alibaba Group
}

\begin{document}

\maketitle

\begin{abstract}
Existing video tokenizers typically use the traditional Variational Autoencoder (VAE) architecture for video compression and reconstruction. However, to achieve good performance, its training process often relies on complex multi-stage training tricks that go beyond basic reconstruction loss and KL regularization. Among these tricks, the most challenging is the precise tuning of adversarial training with additional Generative Adversarial Networks (GANs) in the final stage, which can hinder stable convergence. In contrast to GANs, diffusion models offer more stable training processes and can generate higher-quality results. Inspired by these advantages, we propose \textbf{\ourmethod}, a novel \underline{\textbf{C}}onditioned \underline{\textbf{D}}iffusion-based video \underline{\textbf{T}}okenizer, that replaces the GAN-based decoder with a conditional causal diffusion model. The encoder compresses spatio-temporal information into compact latents, while the decoder reconstructs videos through a reverse diffusion process conditioned on these latents.  During inference, we incorporate a feature cache mechanism to generate videos of arbitrary length while maintaining temporal continuity and adopt sampling acceleration technique to enhance efficiency. Trained using only a basic MSE diffusion loss for reconstruction, along with KL term and LPIPS perceptual loss from scratch, extensive experiments demonstrate that {\ourmethod} achieves state-of-the-art performance in video reconstruction tasks with just a single-step sampling. Even a scaled-down version of {\ourmethod} (3$\times$ inference speedup) still performs comparably with top baselines. Moreover, the latent video generation model trained with {\ourmethod} also exhibits superior performance. The source code and pretrained weights are available at \url{https://github.com/ali-vilab/CDT}.
\end{abstract}

\section{Introduction}
\label{sec:intro}

\begingroup
\renewcommand\thefootnote{}\footnotetext{$\dagger$~Equal contribution~$*$~Work done as a student researcher at Tongyi Lab~$\S$~Corresponding author}%
\addtocounter{footnote}{-1}
\endgroup

Video tokenizers~\citep{wfvae,cosmos,tang2024vidtok,cvvae} bypass the prohibitive computational demands of direct pixel-level manipulation~\citep{sullivan2012overview} by encoding raw videos into compact latent representations. Therefore, video tokenization has become a cornerstone of efficient video generation~\citep{sora,hunyuan,latte,wei2024dreamvideo}. Current video tokenizers are universally grounded in the Variational Autoencoder (VAE) architecture~\citep{vae,vqvae}. Within this architecture, an encoder network, primarily composed of 3D convolutional layers, compresses input videos into compact low-dimensional latent representations. These representations are then upsampled by a deterministic decoder to faithfully reconstruct original videos in the pixel space.

While the training of VAEs inherently relies on a basic reconstruction loss and a KL regularization, these alone are insufficient for developing effective video tokenizers~\citep{opensoraplan,cvvae}. 
Modern tokenizers usually demand multi-stage training~\citep{wfvae,hunyuan} as well as additional training configurations~\citep{cogvideox,opensora,opensoraplan} to function effectively.
These configurations include perceptual loss~\citep{zhang2018unreasonable} for ensuring high-level semantic consistency, 3D GAN  (Generative Adversarial Network) adversarial loss~\citep{esser2021taming} for maintaining temporal consistency and spatial fidelity, and initialization from pre-trained image VAEs to enhance stable convergence~\citep{chen2024od,moviegen,zhou2024allegro}. 
The integration of GANs~\citep{gan}, however, introduces specific challenges. The training process of GANs is inherently unstable and often suffers from issues like mode collapse~\citep{dhariwal2021diffusion}, which necessitates precise and labor-intensive hyperparameter tuning to achieve the desired performance. 
Despite these difficulties, GANs are critical for their ability to significantly improve the overall realism of reconstruction videos. 
This is evident in models like the state-of-the-art HunyuanVideo-VAE~\citep{hunyuan}, which retains adversarial training, even though it has given up using pre-trained image VAEs to initialize parameters, highlighting the indispensable role of GANs in achieving satisfactory results.

\begin{wrapfigure}{r}{0.6\textwidth}
\centering
\begin{minipage}[t]{1\linewidth}
\includegraphics[width=1.0\linewidth,clip,trim=0 0 0 0]{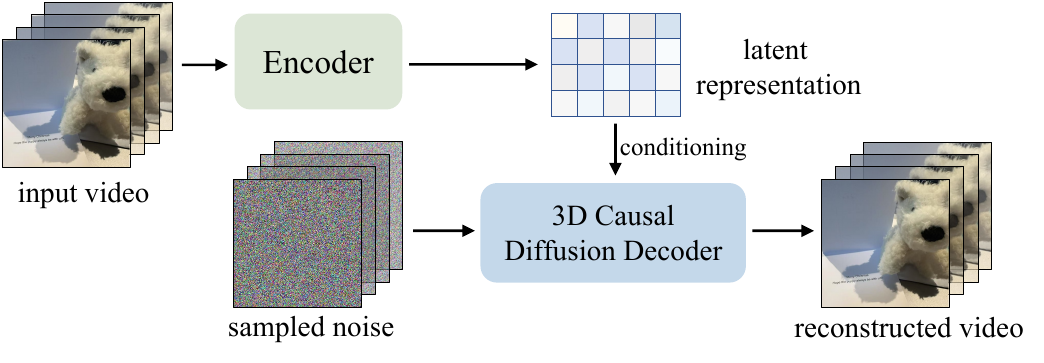}
\caption{\textbf{Overview:}
{\ourmethod} adopts a causal diffusion-based decoder, achieving reconstruction via a reverse diffusion process that is conditioned on the latents extracted by encoder.
}
\label{fig:overview}
\end{minipage}
\end{wrapfigure}

Recently, diffusion models~\citep{ddpm,sgm} have been recognized as a superior alternative to GANs, offering more stable training and generating higher-quality outputs~\citep{dhariwal2021diffusion}. Motivated by this, we set out to explore using diffusion models to overcome the limitations of training decoders with GANs. In this work, as illustrated in Fig.~\ref{fig:overview}, we introduce a novel \underline{\textbf{C}}onditioned \underline{\textbf{D}}iffusion-based video \underline{\textbf{T}}okenizer, entitled \textbf{\ourmethod}. Similar to existing video tokenizers, the encoder in {\ourmethod} compress input videos into compact latent representation. However, {\ourmethod} sets itself apart by using a conditional causal diffusion model for decoding, instead of relying on a decoder trained in the GAN fashion. In the decoding phase, the process begins with noise and iteratively refines it through a reverse diffusion process conditioned on the latent representations from the encoder to reconstruct the videos. Our approach simplifies training compared to existing methods by using only the diffusion model's MSE loss for reconstruction, a KL regularization on the latent space, and an additional LPIPS perceptual loss.
To support arbitrary-length video generation and maintain temporal continuity, a feature caching mechanism is 
incorporated during inference. Moreover, {\ourmethod} leverages Denoising Diffusion Implicit Model (DDIM)~\citep{ddim} for accelerating the diffusion sampling process, enhancing reconstruction efficiency. 
Extensive experiments demonstrate that {\ourmethod} achieves state-of-the-art performance in video reconstruction using just a single sampling step. With reduced model parameters, {\ourmethod} still delivers results comparable to the top baselines. Besides, the latent video generation model built with {\ourmethod} exhibits superior performance. 
\noindent\textbf{Our key contributions are summarized as follows:}
\begin{itemize}[leftmargin=*]
    \item To the best of our knowledge, we introduce the first diffusion-based video tokenizer without complex training tricks (\textit{e.g.,} multi-stage training with GANs) and hope to shed valuable insights for future research.
    \item The decoding process is reformulated as a reverse diffusion process conditioned on the latent representations obtained from the encoder to achieve near-exact reconstruction.
    \item Results show that {\ourmethod} achieves SOTA reconstruction fidelity via single-step sampling, while also yielding superior performance in latent video generation tasks.
\end{itemize}

\section{Preliminaries and Related Works}
\label{sec:related_works}

\subsection{Diffusion Models}
\label{sec:related_works_diffusion_models}

Diffusion models~\citep{ddpm,sgm} have emerged as a powerful framework for generative modeling, especially in the image and video generation tasks~\citep{nichol2022glide,ho2022video,croitoru2023diffusion,esser2023structure,sana}. These models typically comprise two Markov chains: a forward noising process and a reverse learnable denoising process. The forward process gradually adds noise to a clean data $\mathbf{x}_{0}$ over $T$ timesteps according to a pre-defined variance schedule $\{\beta_{t}\}_{t=1}^{T}$. At each timestep $t$, the transition is defined as:
\begin{equation}
\label{eq:ddpm_forward}
q(\mathbf{x}_{t}\vert\mathbf{x}_{t-1})=\mathcal{N}\left(\mathbf{x}_{t};\sqrt{1-\beta_{t}}\mathbf{x}_{t-1},\beta_{t}\mathbf{I}\right).
\end{equation}
Given this formulation, the state $\mathbf{x}_t$ at timestep $t$ can be expressed in a closed-form expression in terms of $\mathbf{x}_0$ and a noise term $\epsilon \sim \mathcal{N}(0, \mathbf{I})$, i.e.,  $\mathbf{x}_{t} = \sqrt{\bar{\alpha}_{t}}\mathbf{x}_{0}+\sqrt{1-\bar{\alpha}_{t}}\epsilon$
where $\alpha_{t}=1-\beta_{t}$ and $\bar{\alpha}_{t}=\Pi_{\tau=1}^{t}\alpha_{\tau}$ is the cumulative product of $\alpha_t$ up to timestep $t$. The reverse process begins with $\mathbf{x}_{T}$ drawn from the prior distribution. It iteratively removes noise predicted by a neural network $\epsilon_{\theta}$ corresponding to the noise injected at the forward timestep, allowing gradually recovering $\mathbf{x}_{0}$. The formulation of the entire denoising process is expressed as $p_\theta(\mathbf{x}_{0:T})=p(\mathbf{x}_T)\prod_{t=1}^Tp_\theta(\mathbf{x}_{t-1}|\mathbf{x}_t)$, where $p_\theta(\mathbf{x}_{t-1}|\mathbf{x}_t)$ is approximated by:
\begin{equation}
\label{eq:ddpm_denoise}
p_{\theta}(\mathbf{x}_{t-1}\vert\mathbf{x}_{t})=\mathcal{N}\left(\mathbf{x}_{t-1};\mu_{\theta}(\mathbf{x}_{t},t), \beta_t \mathbf{I}\right),
\end{equation}
where $\mu_{\theta}(\mathbf{x}_{t},t)=\frac{1}{\sqrt{\alpha_t}} \mathbf{x}_t-\frac{\beta_t}{\sqrt{\alpha_t\left(1-\bar{\alpha}_t\right)}} \epsilon_\theta\left(\mathbf{x}_t, t\right)$ is the predicted posterior mean. The training optimization aims to align $p_\theta(\mathbf{x}_0)$ with the data distribution $q(\mathbf{x}_0)$ and the objective adopts the variational upper bound of the negative log-likelihood:
\begin{equation}
\begin{aligned}
    \mathcal{L}= &\mathbb{E}_q \bigg[-\log p_\theta(\mathbf{x}_0|\mathbf{x}_1)+
    &\sum_{t>1}D_{KL}\left[q(\mathbf{x}_{t-1}|\mathbf{x}_t,\mathbf{x}_0)\parallel p_\theta(\mathbf{x}_{t-1}|\mathbf{x}_t)\right] \bigg] + C.
\end{aligned}
\end{equation}
Denoising Diffusion Probabilistic Model (DDPM)~\citep{ddpm} proposes a simplified objective to train the denoising process parameterized by $\theta$ by directly matching the predicted noise to the actual perturbations added during the forward process, which is formulated as follows:
\begin{equation}
\label{eq:ddpm_objective}
\mathcal{L}=\mathbb{E}_{t,\mathbf{x}_0, \epsilon}\left\|\epsilon-\epsilon_\theta\left(\mathbf{x}_{t}, t\right)\right\|^2.
\end{equation}
Diffusion models are often criticized for their inefficiency in sampling, as they require many iterations to generate high-quality samples. To mitigate this limitation, some techniques~\citep{ddim,dpm_solver,rectified_flow,consistency} have been proposed to enhance sampling efficiency while preserving strong generation quality. Among these, Denoising Diffusion Implicit Model (DDIM)~\citep{ddim} stands out as a representative method. Its primary advantage lies in maintaining the original DDPM training framework while enabling deterministic sampling through a redesigned non-Markovian sampling trajectory. This allows DDIM to achieve comparable generation quality to DDPM with orders-of-magnitude fewer steps.

\subsection{Video Tokenizers}
\label{sec:related_works_video_tokenizers}

Existing video tokenizers are typically built on the variational autoencoder (VAE) architecture~\citep{vae} and can be divided into two categories: discrete and continuous tokenizers. In this paper, we specifically focus on continuous tokenization methods, which map video data into continuous latent representations. The differences between continuous and discrete tokenizers are further discussed in Appendix~\ref{sec:more_related_works}. With our focus established, we proceed to review representative methods. Earlier works~\citep{svd,animatediff,videocrafter1} propose to directly apply image tokenizers to video data via frame-wise compression. However, these methods overlook temporal redundancies across frames. The emergence of OpenAI’s Sora has catalyzed~\citep{sora} some works~\citep{opensora,opensoraplan,wfvae,cosmos,cvvae,cogvideox,hunyuan} aimed at training tokenizers specifically tailored for videos to achieve temporal compression. Among them, OpenSora~\citep{opensora} and OpenSoraPlan~\citep{opensora} stand out as two open-source projects dedicated to re-implementing Sora-like video generative models and they both devise continuous tokenizers for videos to achieve temporal and spatial compression, respectively. CogVideoX~\citep{cogvideox} offers a powerful video VAE with enhanced reconstruction fidelity and, building on this tokenizer, CogVideoX achieves notable text-to-video generation performance. CV-VAE~\citep{cvvae} introduces a latent space regularization to ensure its learned latent space is compatible with that of a given image VAE, allowing efficient video model training using pre-trained text-to-image or video models in a spatio-temporally compressed latent space. More recently, HunyuanVideo~\citep{hunyuan} releases a powerful video VAE that delivers new state-of-the-art performance in both image and video reconstruction.

\section{Methodology}
\label{sec:methodology}

This section is organized as follows:
Sec.~\ref{sec:preliminaries} introduces key notations and formally states the problem; Sec.~\ref{sec:method_formulation} formulates our proposed {\ourmethod}, focusing on its design principles and key ideas; Sec.~\ref{sec:model_instantiation} details the implementation of {\ourmethod}.

\subsection{Notations}
\label{sec:preliminaries}

We follow the causal scenario for video representation, where a video is typically denoted as $\mathbf{V} \in \mathbb{R}^{(1+F) \times H \times W \times 3}$. Here, $1+F$ denotes the total number of frames, each with a height $H$ and width 
$W$ in RGB format. In this setup, the first frame is processed independently as an image for compression purposes, allowing the video tokenizer to effectively handle both image and video tokenization.

In this paper, we focus on the continuous tokenization approach. Our goal is to train a video tokenizer comprising an encoder $\mathcal{E}$, which encodes a video into a compact low-dimensional representation $\mathbf{z}$, and a decoder $\mathcal{D}$, which reconstructs the video from the obtained $\mathbf{z}$. Denoting the reconstructed video as $\hat{\mathbf{V}}$, this process can be formulated as:
\begin{equation}
    \label{seq:vae_formulation}
    \mathbf{z} = \mathcal{E} (\mathbf{V}),~\hat{\mathbf{V}} = \mathcal{D}(\mathbf{z}),\nonumber
\end{equation}
where $\mathbf{z} \in \mathbb{R}^{(1+f) \times h \times w \times c}$. The compression rate is 
defined as $\rho_{t}\times\rho_{s}\times\rho_{s}$, where $\rho_{t}=\frac{F}{f}$ and $\rho_{s}=\frac{H}{h}=\frac{W}{w}$ are the temporal and spatial compression factors, respectively.

\subsection{Method Formulation}
\label{sec:method_formulation}

Our key innovation lies in introducing a novel decoding mechanism, while our encoder $\mathcal{E}$, parameterized by ${\varphi}$, adheres to the design of existing video tokenizers without specific modifications. The encoder compresses a raw input video into a compact latent representation with sufficient expressive power.
To align with the notation in subsequent discussions on diffusion-related decoding, we denote the raw input video as $\mathbf{V}_{0}$. This process is formally expressed as:
\begin{equation}
\label{eq:encoder}
\mathbf{z}=\mathcal{E}_\varphi(\mathbf{V}_{0}).
\end{equation}
The obtained latent representation $\mathbf{z}$ is used differently from existing methods, which typically upsample 
$\mathbf{z}$ directly to the pixel space for video reconstruction. Instead, we use 
$\mathbf{z}$ as the condition for the reverse process in our diffusion-based decoder. In the following, we detail this decoder, implemented within the DDPM~\citep{ddpm} framework. The forward process, starting from the input video $\mathbf{V}_{0}$, follows the noise injection scheme formulated in Eq.~\ref{eq:ddpm_forward}, which progressively corrupts the video over $T$ timesteps. The forward transition at timestep $t$ is defined as:
\begin{equation}
\label{eq:our_forward_process}
q(\mathbf{V}_{t}\vert\mathbf{V}_{t-1})=\mathcal{N}\left(\mathbf{V}_{t};\sqrt{1-\beta_{t}}\mathbf{V}_{t-1},\beta_{t}\mathbf{I}\right),
\end{equation}
where we use the cosine scheduler~\citep{nichol2021improved} for $\beta_{t}$. As for the reverse generative process, the decoder receives the extracted latent $\mathbf{z}$ as a condition. Based on Eq.~\ref{eq:ddpm_denoise}, the conditioned denoising process can be reformulated as:
\begin{equation}
\label{eq:our_reverse_process}
p_{\theta}(\mathbf{V}_{t-1}\vert\mathbf{V}_{t},\mathbf{z})=\mathcal{N}\left(\mathbf{V}_{t-1};\mu_{\theta}(\mathbf{V}_{t},\mathbf{z},t), \beta_t \mathbf{I}\right),
\end{equation}
where $\mu_{\theta}(\mathbf{V}_{t},\mathbf{z},t)$ is further reparameterized by leveraging a noise prediction network $\epsilon_{\theta}$ as follows:
\begin{equation}
\label{eq:network_reparameterize}
\mu_{\theta}(\mathbf{V}_{t},t)=\frac{1}{\sqrt{\alpha_t}} \mathbf{V}_t-\frac{\beta_t}{\sqrt{\alpha_t\left(1-\bar{\alpha}_t\right)}} \epsilon_\theta\left(\mathbf{V}_t,\mathbf{z},t\right).
\end{equation}
\noindent\textbf{Training Objective.} For the diffusion-based decoding process, to estimate the reconstruction ability of the decoder, we introduce a simplified objective to train this reverse generative process as proposed by DDPM~\citep{ddpm}, where the loss function is the mean-squared error between the true noise and the predicted noise at each timestep:
\begin{equation}
\label{eq:our_diffusion_loss_1}
\mathcal{L}_{\text{diffusion}}=\mathbb{E}_{t,\mathbf{V}_0, \epsilon}\left\|\epsilon-\epsilon_\theta\left(\mathbf{V}_{t},\mathbf{z},t\right)\right\|^2.
\end{equation}
The above $\mathcal{L}_{\text{diffusion}}$ can be further reformulated~\citep{salimans2022progressive} as:
\begin{equation}
\label{eq:our_diffusion_loss_2}
\mathcal{L}_{\text{diffusion}}=\mathbb{E}_{t,\mathbf{V}_0, \epsilon}
\frac{\bar{\alpha}_{t}}{1-\bar{\alpha}_{t}}
\left\|\mathbf{V}_{0}-\mathcal{V}_\theta\left(\mathbf{V}_{t},\mathbf{z},t\right)\right\|^2,
\end{equation}
where $\mathcal{V}_{\theta}$ is a learnable network directly predicting the clean data $\mathbf{V}_{0}$. The equivalence between Eq.~\ref{eq:our_diffusion_loss_1} and Eq.~\ref{eq:our_diffusion_loss_2} can be easily derived via $\epsilon_\theta\left(\mathbf{V}_t, \mathbf{z}, t\right)=\frac{\mathbf{V}_t-\sqrt{\bar{\alpha}_{t}} \mathcal{V}_\theta\left(\mathbf{V}_t, \mathbf{z}, t\right)}{\sqrt{1-\bar{\alpha}_{t}}}$. In practice, we adopt Eq.~\ref{eq:our_diffusion_loss_2} to train the reverse diffusion process.

\begin{algorithm}[t!]
\caption{Reconstruction}
\label{alg:reconstruction}
\begin{algorithmic}[1]
    \STATEx {\bfseries Input:} the input video, $\mathbf{V}_{0}$;  
    \STATEx \quad\quad\quad sequence of $N$ time points, $\{\tau_{i}\}_{i=1}^{N}$,
    s.t. $\tau_N=T>\tau_{N-1}>\cdots>\tau_{1}>\tau_0=0$;
    \STATEx \emph{\note{\#~Encode}}
    \STATE $\mathbf{z} = \mathcal{E}_{\varphi}(\mathbf{V}_{0})$
    \STATEx \emph{\note{\#~Sample $\mathbf{V}_T$ from the a standard normal distribution}}
    \STATE $\hat{\mathbf{V}}_{\tau_N}=\mathbf{V}_T\sim\mathcal{N}(\mathbf{0},\mathbf{I})$
    \STATEx \emph{\note{\#~Decode}}
    \FOR{$n=N$ \textbf{to} $1$}
        \STATE $\hat{\mathbf{V}}_{0}^\prime=\mathcal{V}_\theta\left(\hat{\mathbf{V}}_{\tau_n},\mathbf{z},\tau_n\right)$
        \STATE $\epsilon_{\theta}=\frac{\hat{\mathbf{V}}_{\tau_n}-\sqrt{\bar{\alpha}_{\tau_n}} \hat{\mathbf{V}}_{0}^\prime}{\sqrt{1-\bar{\alpha}_{\tau_n}}}$\hspace*{9em}%
        \rlap{\smash{$\left.\begin{array}{@{}l@{}}\\{}\\{}\end{array}\color{black}\right\}%
          \color{black}\begin{tabular}{l}Use DDIM\end{tabular}$}}
        \STATE
        $
\hat{\mathbf{V}}_{\tau_{n-1}}=\sqrt{\bar{\alpha}_{\tau_{n-1}}}\hat{\mathbf{V}}_{0}^\prime +\sqrt{1-\bar{\alpha}_{\tau_{n-1}}}\epsilon_{\theta}
$
    \ENDFOR
    \STATEx {\bfseries Output:} the reconstructed video, $\hat{\mathbf{V}}_0$;
\end{algorithmic}
\end{algorithm}

In addition, a KL regularization on the learned latent space is necessary for facilitating generation. We also introduce a widely-used LPIPS loss~\citep{zhang2018unreasonable} to improve perceptual quality of reconstructed videos. Therefore, the final training objective is given by:
\begin{equation}
\label{eq:final_loss}
\mathcal{L}=\mathcal{L}_{\text{diffusion}}+\lambda\mathcal{L}_{\text{KL}}+\eta\mathcal{L}_{\text{LPIPS}},
\end{equation}
where $\lambda$ and $\eta$ are hyper-parameters. We adopt Eq.~\ref{eq:final_loss} as objective to jointly train the encoder and decoder from scratch. Notably, this objective is simpler than those used in existing models~\citep{cvvae,wfvae,hunyuan} as it excludes adversarial loss, leading to more stable training.

\noindent\textbf{Decoding.} DDPM typically requires hundreds of sampling steps to generate high-quality outputs, with generation time increasing linearly with the number of steps, leading to relatively low sampling efficiency. To address this, we resort to using DDIM~\citep{ddim} sampling method for decoding, which is consistent with the same training approach as DDPM. DDIM can generate high-quality outputs with significantly fewer steps, thereby greatly accelerating our decoding. The sampling formulation of DDIM is presented as follows:
\begin{equation}
\label{eq:ddim_sampling}
\mathbf{V}_{t-1}=\sqrt{\bar{\alpha}_{t-1}} \mathcal{V}_\theta(\mathbf{V}_{t},\mathbf{z},t)+\sqrt{1-\bar{\alpha}_{t-1}}\epsilon_\theta\left(\mathbf{V}_t, \mathbf{z}, t\right).
\end{equation}

The DDIM sampler enables reverse processes with less iteration number than $T$.
Specifically, DDIM considers a diffusion process defined on a subset $\{\mathbf{V}_{\tau_1},\cdots,\mathbf{V}_{\tau_N}\}$, where $\tau$ is an increasing sub-sequence of $[1,\dots,T]$ of length $N$. Here, $N$ is exactly the number of sampling steps required for decoding. $\mathbf{V}_{\tau_{n-1}}$ can be directly derived from $\mathbf{V}_{\tau_{n}}$ via:
\begin{equation}
\label{eq:ddim_sampling_tau}
\mathbf{V}_{\tau_{n-1}}=\sqrt{\bar{\alpha}_{\tau_{n-1}}} \mathcal{V}_\theta(\mathbf{V}_{\tau_{n}},\mathbf{z},\tau_n)+\sqrt{1-\bar{\alpha}_{\tau_{n-1}}}\epsilon_\theta\left(\mathbf{V}_{\tau_n}, \mathbf{z}, \tau_n\right).
\end{equation}

We formalize the reconstruction procedure using DDIM in Alg.~\ref{alg:reconstruction}.
With sufficient training, we observe that even a single DDIM sampling step (\ie, $N=1$) can achieve impressive fidelity with high efficiency. Moreover, increasing the number of sampling steps further enhances reconstruction faithfulness as shown in Sec.~\ref{sec:sensitivity}; however, this comes at the expense of efficiency, highlighting an inherent trade-off.

\subsection{Model Instantiation}
\label{sec:model_instantiation}

\begin{figure}[tb!]
\centering
\includegraphics[width=\linewidth,clip,trim=0 0 0 0]{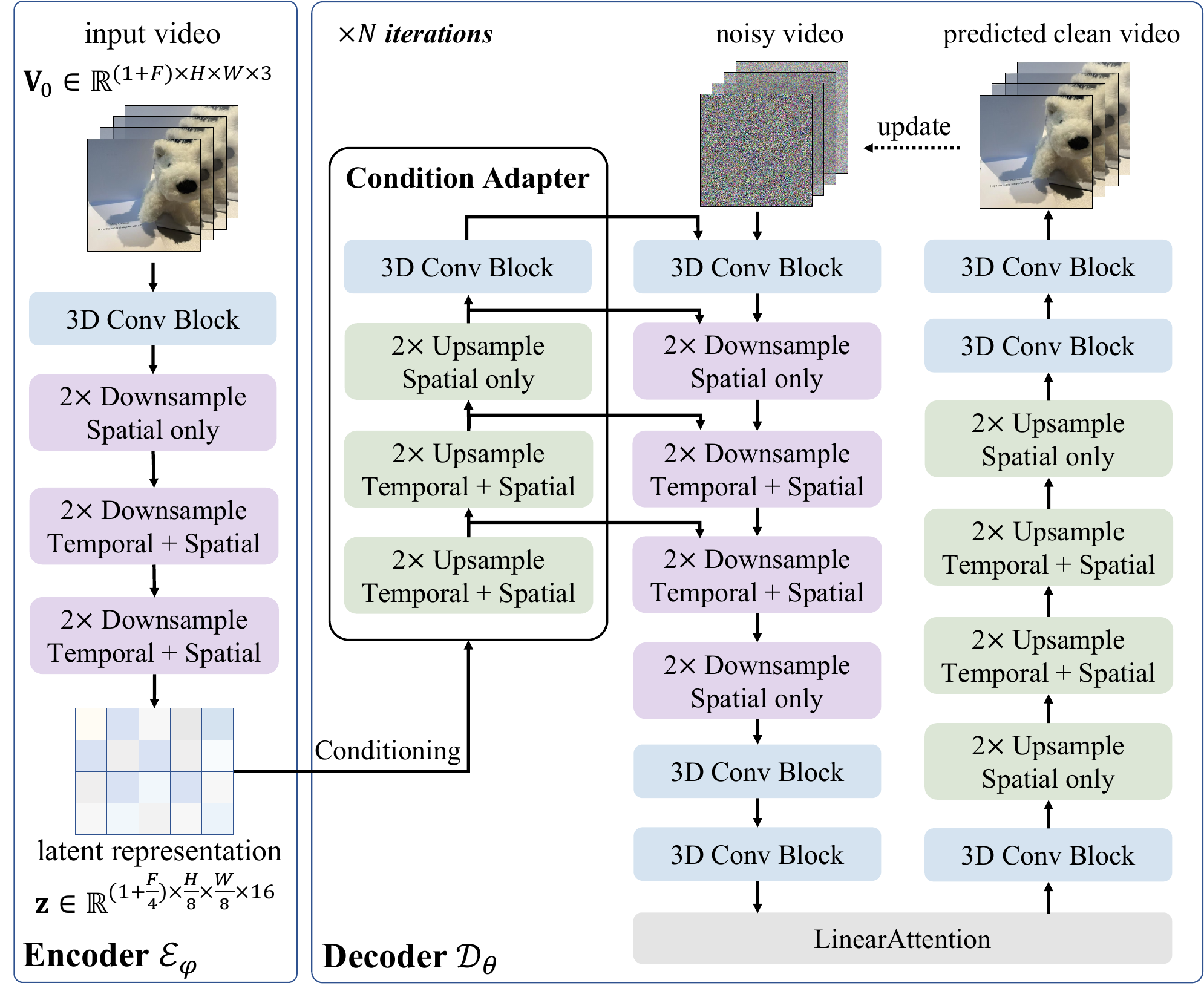}
\caption{The architecture of the proposed {\ourmethod}.}
\label{fig:architecture}
\end{figure}

We highlight the architecture of {\ourmethod} in Fig.~\ref{fig:architecture}. {\ourmethod} is implemented as a causal tokenizer because it is based on 3D causal convolutions~\citep{yu2023language}, ensuring that each frame only accesses information from preceding frames.

\subsubsection{Encoder Implementation}
\label{sec:encoder_implementation}
The encoder is responsible for compressing the input video $\mathbf{V}_{0}\in\mathbb{R}^{(1+F) \times H \times W \times 3}$ into a compact latent representation. The encoder begins with a convolution block composed of 3D causal convolution layers to initially encode the input video. Following this, there is a 3D convolution module that applies $2\times$ compression only on the spatial dimensions. Subsequently, it is followed by two modules, each performing $2\times$ compression on both the temporal and spatial dimensions, ultimately outputting the 16-dimensional latent representation $\mathbf{z}\in\mathbb{R}^{(1+\frac{F}{4}) \times \frac{H}{8} \times \frac{W}{8} \times 16}$. Consistent with the popular compression rate of current mainstream video tokenizers~\citep{opensora,opensoraplan,cogvideox,hunyuan}, our encoder ultimately achieves a compression rate of $4~\times~8~\times~8$.

\subsubsection{Decoder Implementation}
\label{sec:decoder_implementation}
The decoder functions as a conditioned denoising network. We follow the design of the denoising network implemented in DDPM and DDIM, which employs a U-Net-like architecture. The key difference is that they process images using 2D convolution, whereas we handle videos and thus implement a 3D U-Net architecture using 3D causal convolution as our backbone. Specifically, like U-Net, our denoising network is structured into a downsampling stage followed by an upsampling stage, connected by an attention module in between. 
As noted in Sec.~\ref{sec:method_formulation}, our denoising network directly predicts the clean video. Below, we present how to condition the denoising process on the latent representation $\mathbf{z}$.

\noindent\textbf{Condition Injection.} 
We draw inspiration from the method introduced by~\citeauthor{cdc} for incorporating conditions during the denoising process. In our approach, we inject the condition $\mathbf{z}$ at the downsampling stage of the 3D U-Net. We design a module called \textbf{Condition Adapter}, which takes $\mathbf{z}$ as its input. This module consists of four sequentially connected sub-modules, each corresponding in reverse order to the first four downsampling modules of the 3D U-Net. Each sub-module processes the input through a 3D convolution to produce an output that matches the shape of the input of its corresponding module in the downsampling stage. We inject the condition 4 times here in total, and we discuss the effects of varying the number of injections in Sec.~\ref{sec:ablation}.

\noindent\textbf{Solution to Arbitrary-Length Videos Processing.} 
Recall that we follow the notations used in causal scenarios and our proposed {\ourmethod} is implemented as a causal video tokenizer as well. 
To enable memory-efficient encoding and decoding arbitrarily long videos, inspired by audio streaming decoding~\citep{yao2021wenet}, we implement the feature cache mechanism within the 3D causal convolution layer and the Temporal Downsample layer.
For an input video with $1+F$ frames, it is divided into $1+F/4$ chunks, matching the number of latent features. 
The encoding and decoding operation is conducted on each chunk individually, with each chunk handling up to 4 frames to prevent GPU memory overflow. 
To maintain temporal continuity between chunks, frame-level feature caches from the preceding chunk are maintained and integrated into the convolution computations of subsequent chunks. 
In the 3D causal convolution setting, two cached features (convolution kernel size = 3) are maintained, applying zero-padding for the initial chunk and reusing the last two frames from the previous chunk for subsequent caches. 
For scenarios with $2\times$ temporal downsampling (stride = 2), non-initial blocks use a single frame cache to ensure temporal correctness. 
This feature cache mechanism optimizes memory use and preserves video coherence across chunk boundaries, ensuring effective processing for infinite-length videos. Refer to Appendix~\ref{sec:appendix_cache} for a visual aid that clarifies the feature cache mechanism.

\begin{table}[tb!]
\centering
\caption{Reconstruction performance comparison results on COCO-Val (image) and Webvid-Val (video) datasets in terms of PSNR, SSIM and LPIPS metrics. All methods share the same 4~$\times$~8~$\times$~8 compression rate, with their latent representation dimensions being either 4 or 16. The \best{best}, \second{second-best}, \third{third-best} and \textbf{fourth-best} results are highlighted, respectively.
}
\label{tab:reconstruction_performance}
\resizebox{1.0\linewidth}{!}{
\begin{tabular}{c|c|c|c|ccccccccccccc}
   \toprule
   \multirow{3}{*}{\textbf{Model}} & \multirow{3}{*}{\textbf{\makecell{Latent\\Dim.}}} & \multirow{3}{*}{\textbf{\makecell{Comp.\\Rate}}} & \multirow{3}{*}{\textbf{\makecell{Param.\\Count}}} && \multicolumn{3}{c}{\textbf{COCO2017-Val}} && \multicolumn{7}{c}{\textbf{Webvid-Val}} \\ \cline{6-8} \cline{10-16}
   
   &&&&&  \multicolumn{3}{c}{\textbf{Resolution:} \textbf{original}}  &&  \multicolumn{3}{c}{\textbf{Resolution:} \textbf{256~$\times$~256}} && \multicolumn{3}{c}{\textbf{Resolution:} \textbf{720~$\times$~720}} \\ \cline{6-8} \cline{10-12} \cline{14-16}
   & & &  &&  \textbf{PSNR $\uparrow$} & \textbf{SSIM $\uparrow$} &  \textbf{LPIPS $\downarrow$}  &&  \textbf{PSNR $\uparrow$} & \textbf{SSIM $\uparrow$} &  \textbf{LPIPS $\downarrow$} && \textbf{PSNR $\uparrow$} & \textbf{SSIM $\uparrow$} &  \textbf{LPIPS $\downarrow$} \\
   \midrule
   OpenSora-v1.2 & 4 & 4~$\times$~8~$\times$~8 & 393M && 26.85 & 0.7523 & 0.1622 && 29.84 & 0.8289 & 0.1261 && 36.14 & 0.9339 & 0.0711 \\
   OpenSoraPlan-v1.2 & 4 & 4~$\times$~8~$\times$~8 & 239M && 25.93 & 0.7276 & 0.0935 && 29.64 & 0.8372 & 0.0693 && 36.07 & 0.9389 & 0.0421 \\
   WF-VAE & 4 & 4~$\times$~8~$\times$~8 & 147M  && 26.91 & 0.7620 & 0.1473 && 30.30 & 0.8571 & 0.0956 && 37.55 & 0.9533 & 0.0370 \\
   \midrule 
   Cosmos-VAE-CV & 16 & 4~$\times$~8~$\times$~8 & 105M && 27.83 & 0.8060 & 0.1800 && 31.41 & 0.8843 & 0.1168 && 39.79 & 0.9687 & 0.0275 \\
   CVVAE-SD3 & 16 & 4~$\times$~8~$\times$~8 & 182M && 29.48 & \textbf{0.8445} & \third{0.0581} && 33.08 & 0.9157 & \textbf{0.0425} && 40.02 & 0.9713 & 0.0207 \\ 
   CogVideoX-1.5 & 16 & 4~$\times$~8~$\times$~8 & 216M  && \textbf{29.54} & 0.8439 & \textbf{0.0594} && \third{34.67} & \third{0.9390} & \third{0.0338} && \textbf{40.69} & \textbf{0.9766} & \textbf{0.0206} \\
   HunyuanVideo-VAE & 16 & 4~$\times$~8~$\times$~8 & 245M && \second{30.43} & \best{0.8673} & \best{0.0332} && \second{35.15} & \second{0.9397} & \second{0.0197} && \second{42.47} & \second{0.9816} & \best{0.0126} \\
   \midrule
   \ourmethod-S & 16 & 4~$\times$~8~$\times$~8 & 121M && \third{30.11} & \third{0.8569} & 0.0664 && \textbf{34.47} & \textbf{0.9294} & \textbf{0.0425} && \third{42.55} & \third{0.9804} & \third{0.0162} \\
   \ourmethod-B & 16 & 4~$\times$~8~$\times$~8 & 193M && \best{30.48} & \second{0.8653} &  \second{0.0414} && \best{36.38} & \best{0.9542} & \best{0.0195} && \best{42.73} & \best{0.9829} & \second{0.0134} \\
   \bottomrule
\end{tabular}
}
\end{table}

\section{Experiments}
\label{sec:experiments}

This section verifies the effectiveness of our proposed {\ourmethod} through comprehensive comparisons against state-of-the-art baselines. More results can be found in Appendix~\ref{sec:appendix_more_results}.

\subsection{Experimental Setups}
\label{sec:exp_setup}

\noindent\textbf{Datasets for Evaluation.}  
To ensure a fair comparison of reconstruction performance, we follow CVVAE-SD3~\citep{cvvae} and conduct image and video reconstruction on COCO2017-val~\citep{coco} and Webvid-val~\citep{webvid}, respectively, to evaluate the model's ability to capture static and dynamic visual information.
In detail, for image reconstruction setup, we maintain the images at their original resolution. For video reconstruction, we assess the methods at two resolutions by resizing and cropping the videos to 256~$\times$~256 and 720~$\times$~720, extracting 17 frames from each video. 
For the video generation evluation, we use the SkyTimelapse~\citep{skytimelapse} dataset, cropping each video to a resolution of 256~$\times$~256 for training.

\noindent\textbf{Datasets for Training.} We use a hybrid training approach using both image and video data, where YFCC-15M~\citep{yfcc} serves as the image dataset, and OpenVid-1M~\citep{nan2025openvidm} along with a private self-collected dataset are used as the video data. We train two different configurations of {\ourmethod}, denoted as \textbf{{\ourmethod}-S} (Small) and \textbf{{\ourmethod}-B} (Base). {\ourmethod}-S has 121M parameters, while {\ourmethod}-B has 193M parameters. Both models have a latent representation dimension of 16. Further training details are provided in Appendix~\ref{sec:appendix_training_details}.

\noindent\textbf{Baselines.} We compare {\ourmethod} against following state-of-the-art methods, all sharing the same compression rate of 4~$\times$8~$\times$~8 as our {\ourmethod}: OpenSora-v1.2~\citep{opensora}, OpenSoraPlan-v1.2~\citep{opensoraplan}, WF-VAE~\citep{wfvae}, Cosmos-VAE-CV~\citep{cosmos}, CVVAE-SD3~\citep{cvvae}, CogVideoX-1.5~\citep{cogvideox}, HunyuanVideo-VAE~\citep{hunyuan}.

\noindent\textbf{Metrics.} 
For assessing the reconstruction performance of images and videos, we utilize the following metrics: Peak Signal-to-Noise Ratio (\textbf{PSNR}), Structural Similarity Index Measure (\textbf{SSIM}), and Learned Perceptual Image Patch Similarity (\textbf{LPIPS}). For evaluating the video generation quality, we employ the Fr\'echet Video Distance (\textbf{FVD}). Unless otherwise specified, all experiments are run in FP32 precision.

\subsection{Reconstruction}
\label{sec:exp_reconstruction}
We evaluate all methods in terms of image and video reconstruction fidelity and efficiency. Our {\ourmethod} utilizes DDIM for decoding with only one step of sampling in this section.

\noindent\textbf{Reconstruction Performance Comparison.}
Table~\ref{tab:reconstruction_performance} reports the reconstruction results for all methods on images and videos. We first focus on the performance of {\ourmethod}-B. Despite using approximately 21.22\% fewer parameters compared to the top baseline, HunyuanVideo-VAE, CDT-B achieves comparable results in image reconstruction and significantly outperforms HunyuanVideo-VAE in PSNR. Regarding video reconstruction results, {\ourmethod}-B exhibits leading performance against all baseline methods at a resolution of 256~$\times$~256, especially excelling in the PSNR and LPIPS by a notable margin. At the higher resolution of 720~$\times$~720, CDT-B generally maintains its lead, with the exception of a slight lag in the LPIPS metric compared to HunyuanVideo-VAE.

As for {\ourmethod}-S, its parameter count is only larger than that of Cosmos-VAE-CV, yet it still achieves impressive results. In the image reconstruction task, {\ourmethod}-S ranks third in PSNR and SSIM metrics, particularly outperforming CogVideoX-1.5, which has 78.51\% more parameters than {\ourmethod}-S. In the video reconstruction experiments, across both resolutions, {\ourmethod}-S surpasses all other baselines in all three metrics, except for CogVideoX-1.5 and HunyuanVideo-VAE. It is remarkable that even though OpenSora-v1.2 has 3.25 times the number of parameters compared to {\ourmethod}-S, {\ourmethod}-S achieves substantial improvements, with increases of 12.14\% and 13.90\% in PSNR and SSIM, respectively, and a 59.06\% reduction in LPIPS, compared to OpenSora-v1.2. At the 720~$\times$~720 resolution, {\ourmethod}-S is second only to HunyuanVideo-VAE in PSNR and SSIM, with a very small gap. These results further validate the effectiveness of our approach. Moreover, a comparison between {\ourmethod}-B and {\ourmethod}-S reveals performance improvements with increased parameters, highlighting the good scalability of our method.

It's worth mentioning that the encoder of HunyuanVideo-VAE, which is the best-performing method among the baselines, has 100M parameters, whereas our {\ourmethod}-B only has 28M parameters, a 72\% reduction. {\ourmethod}-S has even fewer parameters, only 6 million, reducing the count by 94\%. With such a relatively small number of parameters, the latent representation may be less expressive than that of HunyuanVideo-VAE, which means more information might be lost during the compression process. However, {\ourmethod}-B achieves superior reconstruction fidelity at a 256~$\times$~256 resolution, and {\ourmethod}-S achieves comparable results at a 720~$\times$~720 resolution, further demonstrating the powerful reconstruction capabilities of our causal diffusion-based decoder. Additionally, we also present a qualitative comparison but choose to defer the results to Appendix~\ref{sec:reconstruction_cases}.

\begin{wrapfigure}{l}{0.6\textwidth}
\begin{minipage}[t]{\linewidth}
\centering
\captionof{table}{\label{tab:efficiency}Comparison results of the reconstruction effiency.}
\resizebox{0.9\textwidth}{!}{
    \begin{tabular}{c|c|cc}
    \toprule
    \multirow{2}{*}{\textbf{Precision}}	& \multirow{2}{*}{\textbf{Model}}     &   \multicolumn{2}{c}{\textbf{Time (s)}}   \\  
    \cline{3-4}
    &	& \textbf{256~$\times$~256} &\textbf{720~$\times$~720}    \\ \hline  
    \multirow{3}{*}{FP32}	& HunyuanVideo-VAE    &   0.530  & 6.620     \\  
    & {\ourmethod}-S  &   0.194  &    1.891     \\  
    & {\ourmethod}-B  & 0.610    &  6.408   \\
    \midrule
    \multirow{3}{*}{BF16}	& HunyuanVideo-VAE    &   0.406  & 4.361   \\  
    & {\ourmethod}-S  &   0.132  &  1.034       \\  
    & {\ourmethod}-B  &0.372     &  3.122     \\
    \bottomrule  
    \end{tabular}
}
\end{minipage}%
\end{wrapfigure}

\noindent\textbf{Efficiency Comparison.} 
Aside from fidelity, efficiency is also a crucial factor in evaluating the performance of a video tokenizer reconstruction. Here, we focus on comparing the efficiency with HunyuanVideo-VAE, as its overall fidelity ranks the highest among the baselines. While the primary experiments utilize FP32 precision, BF16 precision is also assessed for a more comprehensive comparison in this study. Table~\ref{tab:efficiency} summarizes the average time cost of reconstructing a single video at both resolutions for our method and HunyuanVideo-VAE, under both precision settings. These experiments were conducted on a single A100 GPU with 80GB of memory. 
In addition to efficiency, we also compare reconstruction fidelity under BF16 precision. The results are provided in Appendix~\ref{sec:bf16}, where our method still achieves superior results.
When using BF16 precision, {\ourmethod}-B consistently outperforms HunyuanVideo-VAE at both resolutions. At the 720~$\times$~720 resolution, our method demonstrates a substantial efficiency advantage, achieving a 28.41\% speedup. Furthermore, {\ourmethod}-B continues to outperform HunyuanVideo-VAE at the 720~$\times$~720 resolution, although there is a slight dip in efficiency at the 256~$\times$~256 resolution. Across both precision settings, {\ourmethod}-B demonstrates an efficiency advantage at the high resolution. This efficiency gain is attributed to the fact that HunyuanVideo-VAE requires tiling to process high-resolution videos, which involves splitting the video into overlapping tiles for separate processing and then merging the outputs to avoid out-of-memory issues. 
In contrast, {\ourmethod} employs a feature cache mechanism as detailed in Sec.~\ref{sec:model_instantiation}, instead of the tiling strategy, thereby avoiding the associated computational overhead. 
Furthermore, recalling the results in Table~\ref{tab:reconstruction_performance}, {\ourmethod}-S achieves PSNR and SSIM results on par with HunyuanVideo-VAE for 720~$\times$~720 resolution videos in FP32 precision, while reducing the time cost by 71.44\%. Additionally, at the 256~$\times$~256 resolution, {\ourmethod}-S also achieves a 63.40\% reduction in time cost compared to HunyuanVideo-VAE. Therefore, {\ourmethod}-S can serve as a powerful tokenizer offering a good trade-off between efficiency and fidelity.

\subsection{Video Generation}
\label{sec:exp_generation}

We further conduct experiments to evaluate whether our {\ourmethod} is effective in the video generation task when combined with the latent diffusion method. Based on the reconstruction results shown in Sec.~\ref{sec:exp_reconstruction}, HunyuanVideo-VAE performs the best among all baselines, so we choose to compare directly with HunyuanVideo-VAE only. We adopt the Latte framework~\citep{latte}, specifically using Latte-XL/2, to train latent video generation models based on the latent spaces learned by {\ourmethod}-B and HunyuanVideo-VAE, respectively, on the SkyTimelapse dataset. Each model is trained for a total of 120k steps using 8~$\times$~A100 GPUs with 80GB memory each.

\begin{wrapfigure}{r}{0.6\textwidth}
\centering
\begin{minipage}[t]{1\linewidth}
    \centering
    \includegraphics[width=0.8\linewidth,clip,trim=5 0 0 0]{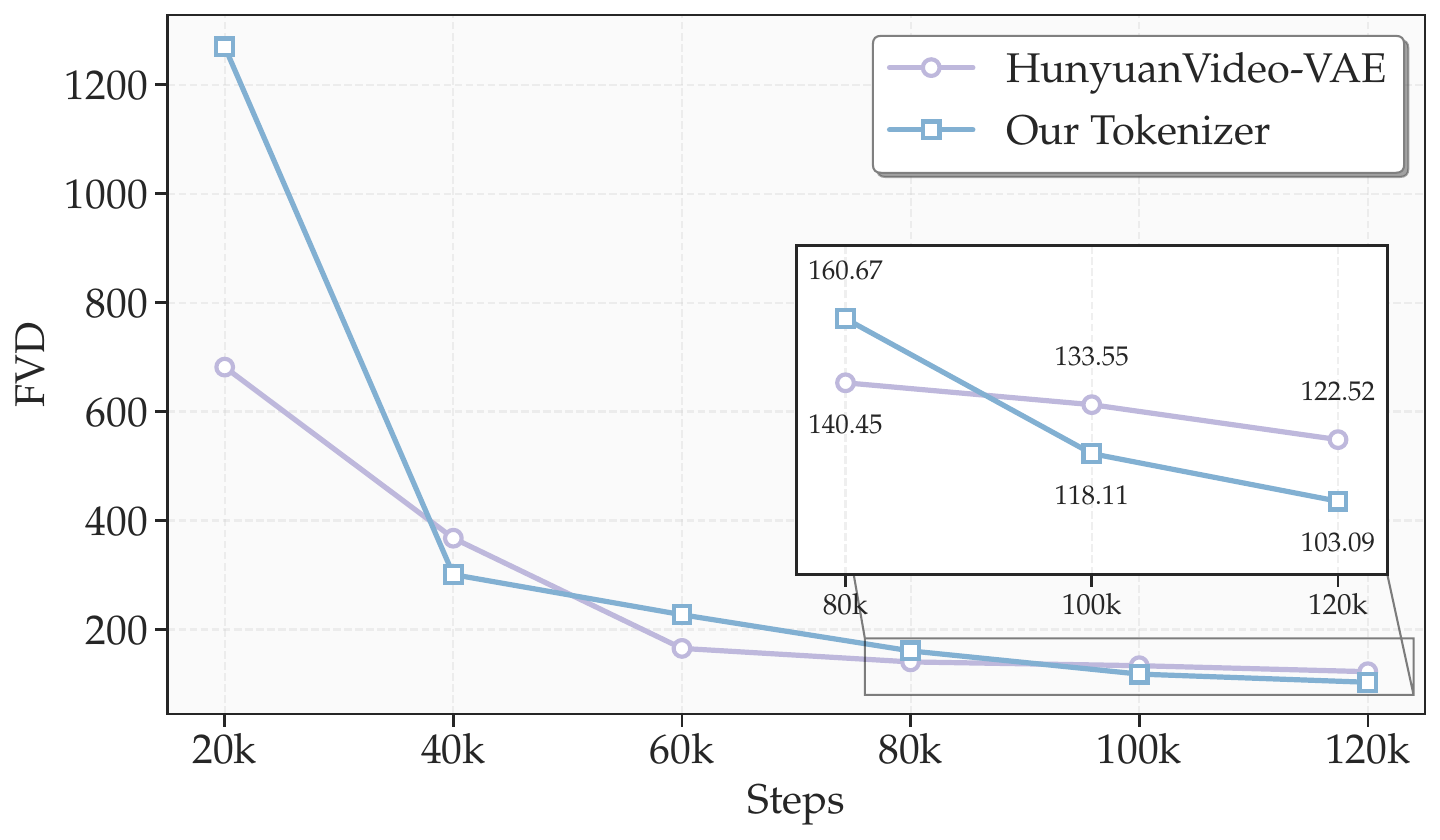}
    \caption{Comparison of FVD trends.}
    \label{fig:fvd_changes}
\end{minipage}
\end{wrapfigure}

The FVD ($\downarrow$) is calculated every 20k steps. We randomly sample 1k real videos from the dataset and fix these samples; each time, we generate 1k videos to calculate the FVD against these fixed real samples. We visualize the change in FVD as the training steps progress in Fig.~\ref{fig:fvd_changes}. Up to the 80k steps, HunyuanVideo-VAE generally outperforms {\ourmethod} in terms of FVD, except for a brief period at the 40k steps where {\ourmethod} performs better. Particularly, at the 20k steps, our FVD is much higher than that of HunyuanVideo-VAE.
Upon analyzing the 16-dimensional latent representations of the SkyTimelapse dataset encoded by both HunyuanVideo-VAE and {\ourmethod},
we find that the values in each dimension of the representation encoded by {\ourmethod} are more concentrated and exhibit lower variance compared to those encoded by HunyuanVideo-VAE.
This indicates that our learned latent space for the SkyTimelapse dataset is more compact, making the denoised latent representation more sensitive to errors. At the 20k steps, the latent diffusion model has not yet converged, leading to inaccurate denoising, which might explain why our method has a much higher FVD at this early stage.

\begin{figure}[tb!]
\centering
\includegraphics[width=0.7\linewidth,clip,trim=0 0 0 0]{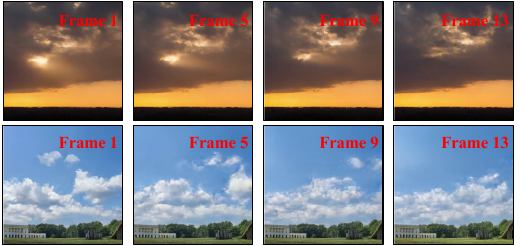}
\caption{Examples of videos generated by Latte using {\ourmethod}.}
\label{fig:generated_video_demo}
\end{figure}

As the steps increase, we see that from the 80k to 100k steps, the FVD of HunyuanVideo-VAE decreases by 4.91\%, while during the same period, {\ourmethod} manages a 26.49\% reduction. At the 100k steps, our FVD is 11.56\% lower than that of HunyuanVideo-VAE. Continuing training for 20k more steps to 120k, we find that the FVD of the {\ourmethod}-based model remains 15.86\% lower than that of HunyuanVideo-{\ourmethod}-based. This indicates that, in the later stages of training, the {\ourmethod}-based model generates videos of higher quality than those based on HunyuanVideo-VAE, demonstrating that our method is a powerful tokenizer for video generation. In Fig.~\ref{fig:generated_video_demo}, we present two videos generated by the {\ourmethod}-based model trained for 120k steps, which appear quite realistic. Particularly, the first row features a video where clouds gradually obscure the sun, causing observable changes in the transmitted light, effectively demonstrating {\ourmethod}'s ability to generate videos that reflect certain laws of physics.

\begin{table}[tb!]
\centering
\caption{Ablation study on videos at 256~$\times$~256 resolution.}
\label{tab:ablation}
\begin{tabular}{l|ccccc}
    \toprule
    \multirow{2}{*}{\textbf{Model Configuration}}  &&  \multicolumn{3}{c}{\textbf{Webvid-Val}} \\    
    \cline{3-5} 
    &&  \textbf{PSNR $\uparrow$} & \textbf{SSIM $\uparrow$} &  \textbf{LPIPS $\downarrow$}\\
    \midrule
    \ourmethod-S (default model configuration) && 32.49 & 0.9060 & 0.0539 \\
    \midrule
    discard $\mathcal{L}_{\text{LPIPS}}$ for training && 31.29 & 0.8875 & 0.1050 \\
    \midrule
    only inject condition once && 31.09 & 0.8916 & 0.0642 \\
    only inject condition twice  && 32.08 & 0.9055 & 0.0592 \\
     only inject condition three times && 32.23 & 0.9057 &  0.0536 \\
    \bottomrule
\end{tabular}
\end{table}

\subsection{Ablation Studies}
\label{sec:ablation}

In this study, we investigate the effects of different components on the final reconstruction performance of our proposed {\ourmethod}. Due to high computational and time requirements, we use the smaller {\ourmethod}-S model for analysis. Each model, including all variants and the full version, is trained for 140k steps using 4~$\times$~80G A100 GPUs to ensure fair comparison.
We evaluate the models on Webvid-Val at 17~$\times$~256~$\times$~256 resolution, presenting results in Table~\ref{tab:ablation}. 

\noindent\textbf{Effect of the} $\mathcal{L}_{\text{LPIPS}}$\textbf{.} As noted in Eq.~\ref{eq:final_loss}, our final training objective consists of three components: $\mathcal{L}_{\text{diffusion}}$, $\mathcal{L}_{\text{KL}}$, and an optional term $\mathcal{L}_{\text{LPIPS}}$. Here, we investigate the impact of $\mathcal{L}_{\text{LPIPS}}$ on performance. When this component is omitted, we observe a degradation in performance across all our evaluation metrics: PSNR, SSIM, and LPIPS. The decrease in the LPIPS score is particularly pronounced and expected, as this metric is directly related to the perceptual similarity that $\mathcal{L}_{\text{LPIPS}}$ aims to improve. Besides, the removal also negatively affects PSNR and SSIM scores, suggesting that $\mathcal{L}_{\text{LPIPS}}$ contributes to enhancing not only perceptual quality but also the fidelity and structural similarity of the generated outputs.

\noindent\textbf{Effect of the Condition Injection Times.} 
As mentioned in Sec.~\ref{sec:model_instantiation}, in our implementation, we choose to inject the encoded latent representation into the first four modules of the downsampling stage of the denoising network, which adopts a 3D U-Net architecture, to condition the denoising process. Here, we experiment with varying the number of condition injections, comparing injections into just the first one, the first two, the first three, and the first four modules. As shown in Table~\ref{tab:ablation}, performance generally improves with more injections across all metrics. More injections typically enhance performance by providing richer conditioning information at multiple stages, thereby enabling the network to perform more accurate denoising. We do not explore more than four injections because it would introduce additional parameters, and since four injections already yield satisfactory results according to our experiments, further increasing the number of injections may not be necessary.

\subsection{Hyper-parameters Sensitivity}
\label{sec:sensitivity}

This study investigates the sensitivity of our method to these hyper-parameters: \emph{i)} the timestep number for diffusion training and \emph{ii)} the sampling steps for DDIM method used in decoding. Similar to Sec.~\ref{sec:ablation}, we evaluate models based on {\ourmethod}-S on videos at the 17~$\times$~256~$\times$~256 resolution. 
Additionally, we also discuss the effect of the latent dimension separately in Appendix~\ref{sec:appendix_latent_dim}.

\noindent\textbf{Effects of Timestep Number on Diffusion Training.} We set the number of timesteps for diffusion to 8192, whereas most works typically adopt around 1000 timesteps~\citep{ddpm,dhariwal2021diffusion,li2023t2t}. Here, we compare the effects of using 8192 versus 1024 steps, with each model trained on 4~$\times$~A100 GPUs. Table~\ref{tab:diffusion_step} presents the reconstruction results on Webvid-Val after different numbers of training steps. 
We can see that the performance of both models generally improves across all metrics as the number of training steps increases. 
A closer inspection reveals that after 70k training steps, the 8192-step model achieves superior SSIM and LPIPS values compared to the 1024-step model. However, the PSNR metric is better for the 1024-step model. This situation remains the same at 80k steps, with the 1024-step model still outperforming in PSNR while lagging in SSIM and LPIPS. By 90k steps, the 8192-step model surpasses the 1024-step model across all metrics. 
To ensure that the performance advantage of the model with more timesteps at 90k steps is not just temporary, we conduct additional training for 60k steps and retest at 150k steps. 
The results show that the model trained with more timesteps consistently outperforms the 1024-step model in the later stages of training.

\begin{wrapfigure}{l}{0.6\textwidth}
\begin{minipage}[t]{\linewidth}
\centering
\captionof{table}{\label{tab:diffusion_step}Video reconstruction results with 8192 \emph{vs.} 1024 diffusion timesteps, with \textbf{best} results highlighted per row.}
\resizebox{0.99\textwidth}{!}{
\begin{tabular}{c|ccccccc}
   \toprule
   \multirow{2}{*}{\textbf{Step}} &  \multicolumn{3}{c}{\textbf{1024 diffusion steps}} && \multicolumn{3}{c}{\textbf{8192 diffusion steps}}\\
   \cline{2-4} \cline{6-8}
    &  \textbf{PSNR $\uparrow$} & \textbf{SSIM $\uparrow$} &  \textbf{LPIPS $\downarrow$} && \textbf{PSNR $\uparrow$} & \textbf{SSIM $\uparrow$} &  \textbf{LPIPS $\downarrow$} \\
   \midrule

   \textsf{@}70k & \textbf{30.63} & {0.8613} & {0.0800} &&  30.54  & \textbf{0.8780}  & \textbf{0.0675}  \\ \hline
   \textsf{@}80k & \textbf{31.28} & 0.8843  & 0.0690 && 30.88  & \textbf{0.8864}  & \textbf{0.0666}  \\ \hline
   \textsf{@}90k & 31.54 & 0.8897 & 0.0647 && \textbf{31.67} & \textbf{0.8954}  & \textbf{0.0611}   \\ \hline
   \textsf{@}150k & 31.69 & 0.8779 & 0.0536 && \textbf{32.08} & \textbf{0.9036} & \textbf{0.0520}   \\
   \bottomrule
\end{tabular}
}
\end{minipage}%
\end{wrapfigure}

The phenomenon arises from the interplay between timestep granularity and training dynamics in diffusion models, which is rooted in the assumption that time intervals are sufficiently small for more accurate noise estimation in diffusion models. Models with fewer timesteps partition the diffusion process into coarser intervals, simplifying the learning of broad noise-reversal patterns and enabling faster initial convergence, which boosts early performance metrics. In contrast, models with more timesteps discretize the process into finer intervals, theoretically allowing for more precise noise estimation and higher-quality outputs. Yet, this granularity demands extended training to discern subtle inter-step dependencies and optimize the increased complexity of transitions, causing them to initially lag. Over time, as training progresses, the finer temporal resolution of high-timestep models enables superior noise modeling and detail synthesis, ultimately surpassing their low-timestep counterparts. This highlights a trade-off between training efficiency and ultimate performance.

\begin{wrapfigure}{l}{0.6\textwidth}
\begin{minipage}[t]{\linewidth}
\centering
\captionof{table}{\label{tab:sampling_step}Reconstruction fidelity and efficiency with varying DDIM sampling steps on a A100 GPU with 80G memory.}
\resizebox{0.99\textwidth}{!}{
\begin{tabular}{c|ccc|c}
   \toprule
   \multirow{2}{*}{\textbf{Model}} &  \multicolumn{4}{c}{\textbf{Webvid-Val}} \\
   \cline{2-5}
    &  \textbf{PSNR $\uparrow$} & \textbf{SSIM $\uparrow$} &  \textbf{LPIPS $\downarrow$} & \textbf{Time} \\
   \midrule
   CogVideoX-1.5 & {34.67} & {0.9390} & {0.0338} & 0.695s \\
   HunyuanVideo-VAE & {35.15} & {0.9397} & {0.0197} & 0.530s \\
   \midrule
   \ourmethod-S (1-step DDIM) & 34.47 & 0.9294 & 0.0425 & 0.194s \\
   \ourmethod-S (2-step DDIM) & 34.53 & 0.9311 & 0.0422 & 0.353s \\
   \ourmethod-S (3-step DDIM) & 34.72 & 0.9337 & 0.0416 & 0.513s \\
   \bottomrule
\end{tabular}
}
\end{minipage}%
\end{wrapfigure}

\noindent\textbf{Effect of Sampling Steps for Decoding.} In Sec.~\ref{sec:exp_reconstruction}, we present the {\ourmethod} results obtained using one-step DDIM~\citep{ddim} sampling during decoding. The original DDIM paper indicates that increasing sampling steps can enhance generation quality. Here, we further investigate the impact of increasing DDIM sampling steps on the performance of {\ourmethod}. Since decoding time scales linearly with the number of sampling steps, increased steps will reduce the efficiency of our video tokenizer. Therefore, we need to balance generation quality and efficiency well. We use the time cost of HunyuanVideo-VAE as a reference for the maximum acceptable time cost, given its superior fidelity among the baselines. With three sampling steps, our method's time cost comes very close to that of HunyuanVideo-VAE, although it is still slightly less. However, with four steps, the time cost would exceed that of HunyuanVideo-VAE. 
Thus, we limit the sampling steps to $\{1, 2, 3\}$. The reconstruction fidelity and time costs on Webvid-Val are summarized in Table~\ref{tab:sampling_step}. This table also includes results from CogVideoX-1.5 and HunyuanVideo-VAE, the two best-performing baselines, for comparison. As shown in Table~\ref{tab:sampling_step}, increasing the DDIM sampling steps improves the reconstruction fidelity of {\ourmethod}-S across all three metrics. 
Specifically, with only one sampling step, {\ourmethod}-S is inferior to CogVideoX-1.5 in PSNR. However, with three steps, {\ourmethod}-S surpasses CogVideoX-1.5 in PSNR. 
We do not use {\ourmethod}-B with increased sampling steps in this evaluation because its time cost with a single-step sampling is already very close to that of HunyuanVideo-VAE. Increasing the steps would make its time cost significantly higher than HunyuanVideo-VAE's.
In practical applications, within an acceptable time cost, we can increase sampling steps to enhance reconstruction quality, at the cost of some efficiency.

\section{Conclusion}
\label{sec:conclusion}
We propose {\ourmethod}, a novel diffusion-based video tokenizer, that adopts a 3D conditional causal diffusion model as the decoder. 
Without overcomplicated training tricks such as multi-stage training with GANs, we elegantly use a simple MSE diffusion loss to achieve high-quality spatio-temporal compression and reconstruction.
Furthermore, to support video reconstruction of arbitrary lengths,
a feature cache mechanism is employed, alongside sampling acceleration technique to enhance decoding efficiency. {\ourmethod} achieves state-of-the-art performance in video reconstruction with single-step sampling, achieves competitive results even with a smaller model, and excels in latent video generation.

\bibliographystyle{plainnat}
\bibliography{reference.bib}

\onecolumn
\appendix

\clearpage
\setcounter{table}{0}
\setcounter{section}{0}
\begin{center}
    {\Huge
    \textbf{Appendix of {\ourmethod}}}\\
    \vspace{1.0em}
\end{center}

\setcounter{table}{0}
\setcounter{figure}{0}
\setcounter{algorithm}{0}

\renewcommand{\thetable}{S\arabic{table}}
\renewcommand{\thefigure}{S\arabic{figure}}
\renewcommand{\thealgorithm}{S\arabic{algorithm}}

\section{More Related Works Discussion on Video Tokenization}
\label{sec:more_related_works}

Existing video tokenizers are typically built on the variational autoencoder (VAE) architecture~\citep{vae} and can be divided into two categories: discrete and continuous tokenizers. Discrete tokenizers~\citep{videogpt,yu2024language,omnitokenizer} adapt the quantization techniques from discrete image tokenization~\citep{vqvae} by mapping video frames into a latent space and quantizing these representations via selecting the nearest vectors from a vector codebook. They are commonly used for autoregressive generation, as they can effectively mitigate error accumulation. In contrast, continuous tokenizers~\citep{cvvae,wfvae,hunyuan} encode videos into low-dimensional latent representations without quantization and generally exhibit superior reconstruction capabilities than discrete methods. In particular, Latent Video Diffusion Models (LVDMs)~\citep{cogvideox,opensora,opensoraplan} successfully integrate continuous tokenizers with latent diffusion methods~\citep{latentdiffusion}, yielding impressive video generation performance.

\section{Supplementary Illustration of the Feature Cache Mechanism}
\label{sec:appendix_cache}

\begin{figure}[h!]
\centering
\includegraphics[width=1.0\linewidth,clip,trim=0 0 0 0]{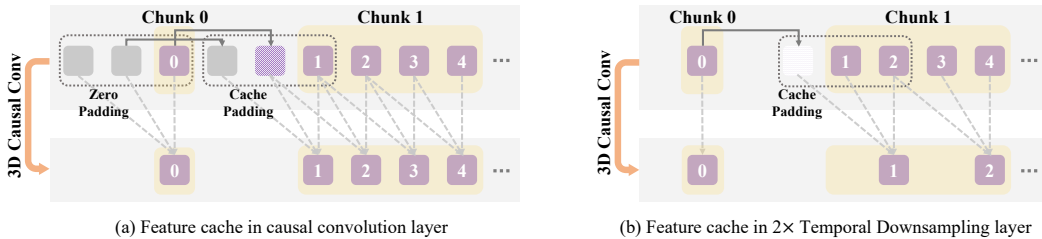}
\vspace{-15pt}
\caption{Illustration of the architecture and operational flow of the feature cache mechanism.}
\label{fig:feature_cache}
\vspace{-15pt}
\end{figure}

This section provides an illustration of the feature cache mechanism in Fig.~\ref{fig:feature_cache}, which is intended to facilitate a deeper understanding of its architecture and functionality, serving as a visual aid to complement the written explanation.

\section{More Training Details of {\ourmethod}}
\label{sec:appendix_training_details}
The training of {\ourmethod} consists of two stages, primarily differing in the resolution and frame count of the video data. In the first stage, the model is trained on low-resolution videos (256~$\times$~256) with a small number of frames (9 or 17) to accelerate convergence, without introducing the LPIPS perceptual loss. In the second stage, training progresses to videos with higher resolutions (\eg, 480~$\times$~480 and 512~$\times$~512) and a larger number of frames (\eg, 25 and 33), at which point the LPIPS perceptual loss is incorporated with a weighting coefficient of 0.01. The model is trained for a total of 400K steps on 16 Nvidia A100 GPUs. The frame rate (FPS) of all training videos is randomly set between 16 and 60 to facilitate learning across a variety of motion speeds. With this training setup, we obtain the primary configuration of our model, denoted as \textbf{{\ourmethod}-B} (Base), which has 193 million parameters and a latent representation dimension of 16. In addition to the base model, we also train a smaller model, \textbf{{\ourmethod}-S} (Small), with reduced parameters and fewer training steps. {\ourmethod}-S has 121 million parameters, while still maintaining a latent representation dimension of 16.

\section{More Experimental Results}
\label{sec:appendix_more_results}

\subsection{Reconstruction Results under BF16 Precision}
\label{sec:bf16}
We present the comparison results of reconstruction performance for various methods on images and videos under BF16 precision in Table~\ref{tab:reconstruction_performance_bf16}. It can be observed that performance generally exhibits a slight decline across methods when precision is lowered. Our {\ourmethod}-B still achieves the best performance across all metrics for video reconstruction at a 256~$\times$~256 resolution. For video reconstruction at a 720~$\times$~720 resolution, it is comparable to HunyuanVideo-VAE. A scaled-down version of our model, i.e., {\ourmethod}-S, also still demonstrates impressive results at this lower precision. Recalling the comparison provided in Table~\ref{tab:efficiency} of the main text under BF16 precision, our method offers a significant efficiency advantage over HunyuanVideo-VAE, while still maintaining a good balance in reconstruction fidelity.

\begin{table}[tb!]
\centering
\caption{Reconstruction fidelity comparison results in BF16 precision on COCO-Val and Webvid-Val  datasets in terms of PSNR, SSIM and LPIPS metrics. All methods share the same 4~$\times$~8~$\times$~8 compression rate, with their latent representation dimensions being either 4 or 16. The \best{best}, \second{second-best}, \third{third-best} and \textbf{fourth-best} results are highlighted, respectively.
}
\label{tab:reconstruction_performance_bf16}
\resizebox{0.96\linewidth}{!}{
\begin{tabular}{c|c|c|c|ccccccccccccc}
   \toprule
   \multirow{3}{*}{\textbf{Model}} & \multirow{3}{*}{\textbf{\makecell{Latent\\Dim.}}} & \multirow{3}{*}{\textbf{\makecell{Comp.\\Rate}}} & \multirow{3}{*}{\textbf{\makecell{Param.\\Count}}} && \multicolumn{3}{c}{\textbf{COCO2017-Val}} && \multicolumn{7}{c}{\textbf{Webvid-Val}} \\ \cline{6-8} \cline{10-16}
   
   &&&&&  \multicolumn{3}{c}{\textbf{Resolution:} \textbf{original}}  &&  \multicolumn{3}{c}{\textbf{Resolution:} \textbf{256~$\times$~256}} && \multicolumn{3}{c}{\textbf{Resolution:} \textbf{720~$\times$~720}} \\ \cline{6-8} \cline{10-12} \cline{14-16}
   & & &  &&  \textbf{PSNR $\uparrow$} & \textbf{SSIM $\uparrow$} &  \textbf{LPIPS $\downarrow$}  &&  \textbf{PSNR $\uparrow$} & \textbf{SSIM $\uparrow$} &  \textbf{LPIPS $\downarrow$} && \textbf{PSNR $\uparrow$} & \textbf{SSIM $\uparrow$} &  \textbf{LPIPS $\downarrow$} \\
   \midrule
   OpenSora-v1.2 & 4 & 4~$\times$~8~$\times$~8 & 393M && 26.83 & 0.7519 & 0.1624 && 29.82 & 0.8283 & 0.1263 && 36.07  & 0.9335 & 0.0712 \\
   OpenSoraPlan-v1.2 & 4 & 4~$\times$~8~$\times$~8 & 239M && 25.93 & 0.7274 & 0.0936 && 29.64 & 0.8367 & 0.0694 && 36.08 & 0.9385 & 0.0421 \\
   WF-VAE & 4 & 4~$\times$~8~$\times$~8 & 147M && 26.89 & 0.7608 & 0.1467 && 30.26 & 0.8556 & 0.0955 && 37.37 & 0.9513 &  0.0372 \\
   \midrule 
   Cosmos-VAE-CV & 16 & 4~$\times$~8~$\times$~8 & 105M && 27.83 & 0.8059 & 0.1801 && 31.40 & 0.8842 & 0.1168  && 39.79 & 0.9686 & 0.0275 \\
   CVVAE-SD3 & 16 & 4~$\times$~8~$\times$~8 & 182M && 29.46 & \textbf{0.8443} & \third{0.0580} && 33.04 & \textbf{0.9151} & 0.0425 && 40.01 & \textbf{0.9703} & 0.0207  \\
   CogVideoX-1.5 & 16 & 4~$\times$~8~$\times$~8 & 216M && \textbf{29.52} & 0.8440 & \textbf{0.0590} && \third{34.65} & \second{0.9393} & \third{0.0335} && \textbf{40.66} & \third{0.9769} & \textbf{0.0204} \\
   HunyuanVideo-VAE & 16 & 4~$\times$~8~$\times$~8 & 245M && \second{30.42} & \best{0.8670} & \best{0.0334} && \second{35.09}  & \second{0.9393} & \second{0.0198} && \best{42.34} & \best{0.9810} & \best{0.0128} \\   %
   \midrule 
   \ourmethod-S & 16 & 4~$\times$~8~$\times$~8 & 121M && \third{30.05} & \third{0.8555} & 0.0653 && \textbf{34.32} & \third{0.9276}  & \textbf{0.0420}  && \third{42.09} & \second{0.9784} & \third{0.0160} \\
   \ourmethod-B & 16 & 4~$\times$~8~$\times$~8 & 193M && \best{30.45} & \second{0.8646} & \second{0.0414} && \best{36.18} & \best{0.9526} & \best{0.0196} && \second{42.29} & \best{0.9810} & \second{0.0137}  \\
   \bottomrule
\end{tabular}
}
\end{table}

\subsection{Reconstruction Results: A Qualitative Perspective}
\label{sec:reconstruction_cases}
\begin{figure}[tb!]
\centering
\includegraphics[width=1.0\linewidth,clip,trim=0 0 0 0]{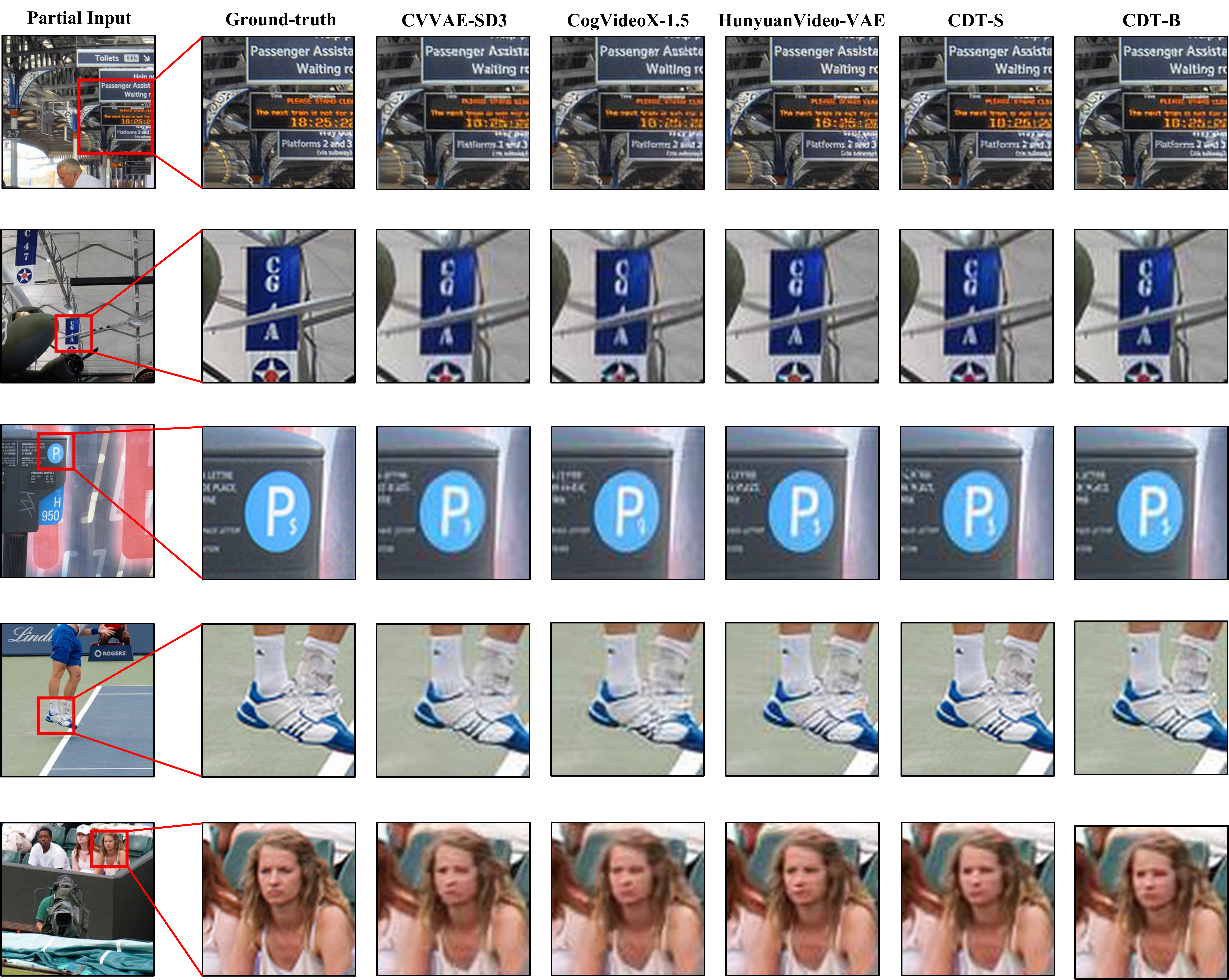}
\caption{Examples of image reconstruction results using different methods, highlighting the superior local detail reconstruction by {\ourmethod}.}
\label{fig:rec_image}
\end{figure}

\begin{figure}[tb!]
\centering
\includegraphics[width=1.0\linewidth,clip,trim=0 0 0 0]{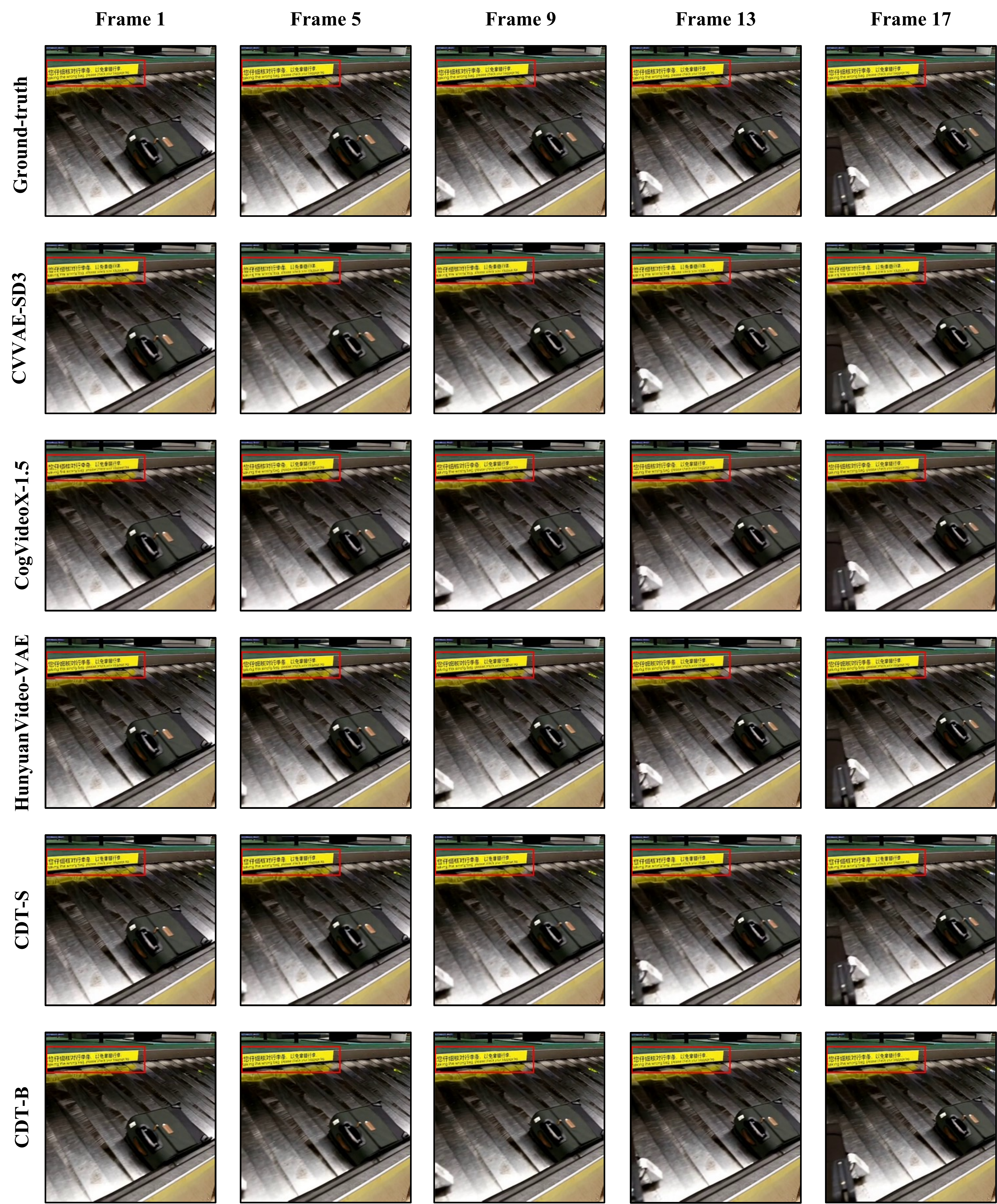}
\caption{Examples of video reconstruction results demonstrating the superior ability of our method to consistently reconstruct small text details.}
\label{fig:rec_video_case1}
\end{figure}

\begin{figure}[tb!]
\centering
\includegraphics[width=1.0\linewidth,clip,trim=0 0 0 0]{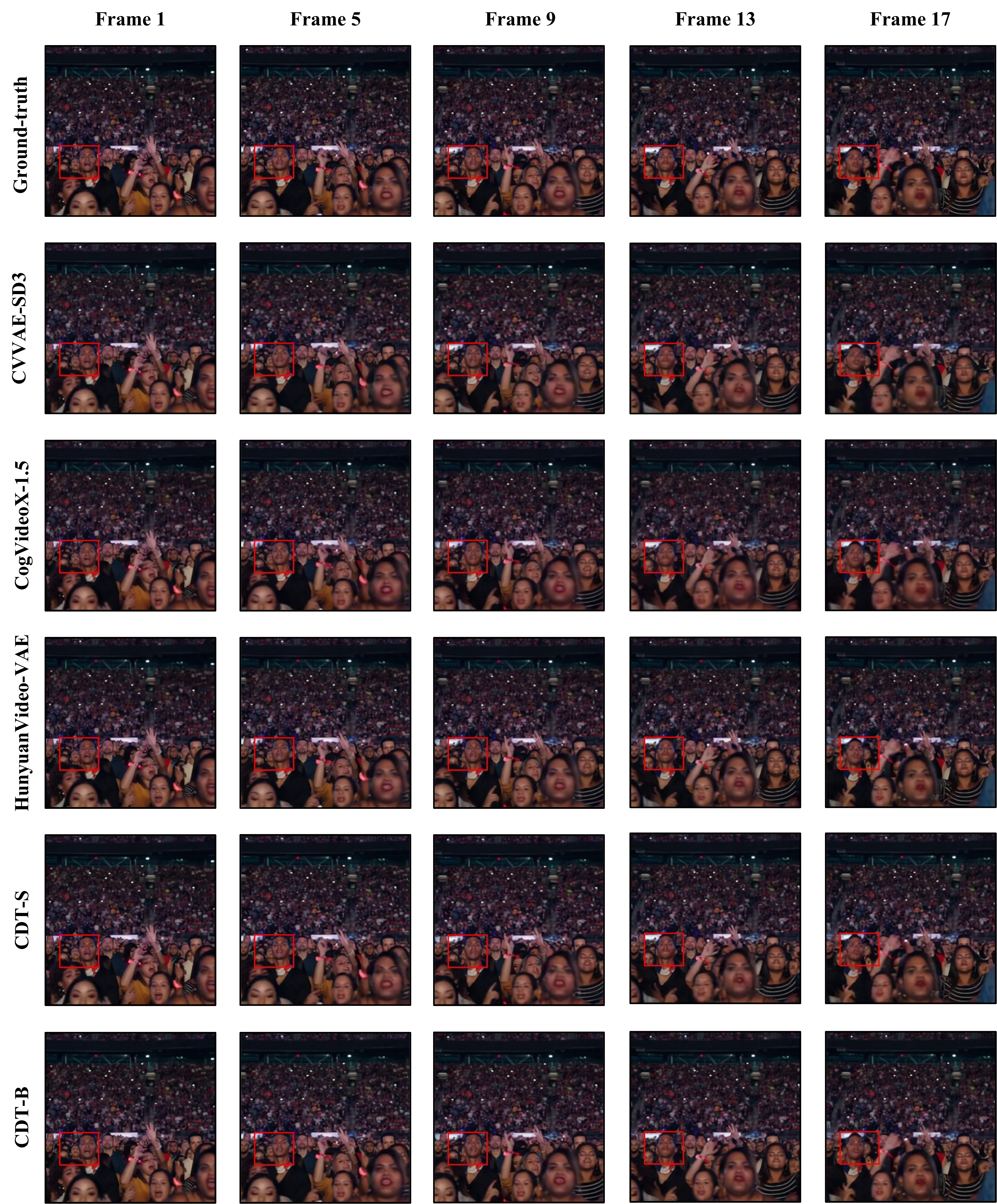}
\caption{Examples of video reconstruction results in a crowded scene, showcasing our method's enhanced ability to consistently reconstruct dynamic facial details.}
\label{fig:rec_video_case2}
\end{figure}

For a qualitative evaluation, we illustrate some images and videos reconstructed by {\ourmethod}-S and {\ourmethod}-B, as well as by other selected baseline tokenizers in this section. We select CVVAE-SD3, CogVideoX-1.5, and HunyuanVideo-VAE for comparison here because these three have the top overall performance among the baselines. Fig.~\ref{fig:rec_image} shows several groups of images, each comparing the reconstruction results of different methods on the same image, where it can be observed that our {\ourmethod}-B method is able to better reconstruct local details. In Fig.~\ref{fig:rec_video_case1} and Fig.~\ref{fig:rec_video_case2}, we present the reconstruction results of one video each. In the example from Fig.~\ref{fig:rec_video_case1}, our method consistently provides better reconstruction of small text in the video. In the example from Fig.~\ref{fig:rec_video_case2}, in a video with densely moving people, we can reconstruct dynamic facial details more consistently than other methods. The {\ourmethod}-S, with even fewer parameters, also achieves good reconstruction results and realism.

\subsection{More Generated Videos by the Model trained in Sec.~\ref{sec:exp_generation}}
\label{sec:more_generated_videos}

\begin{figure}[tb!]
\centering
\includegraphics[width=0.6\linewidth,clip,trim=0 0 0 0]{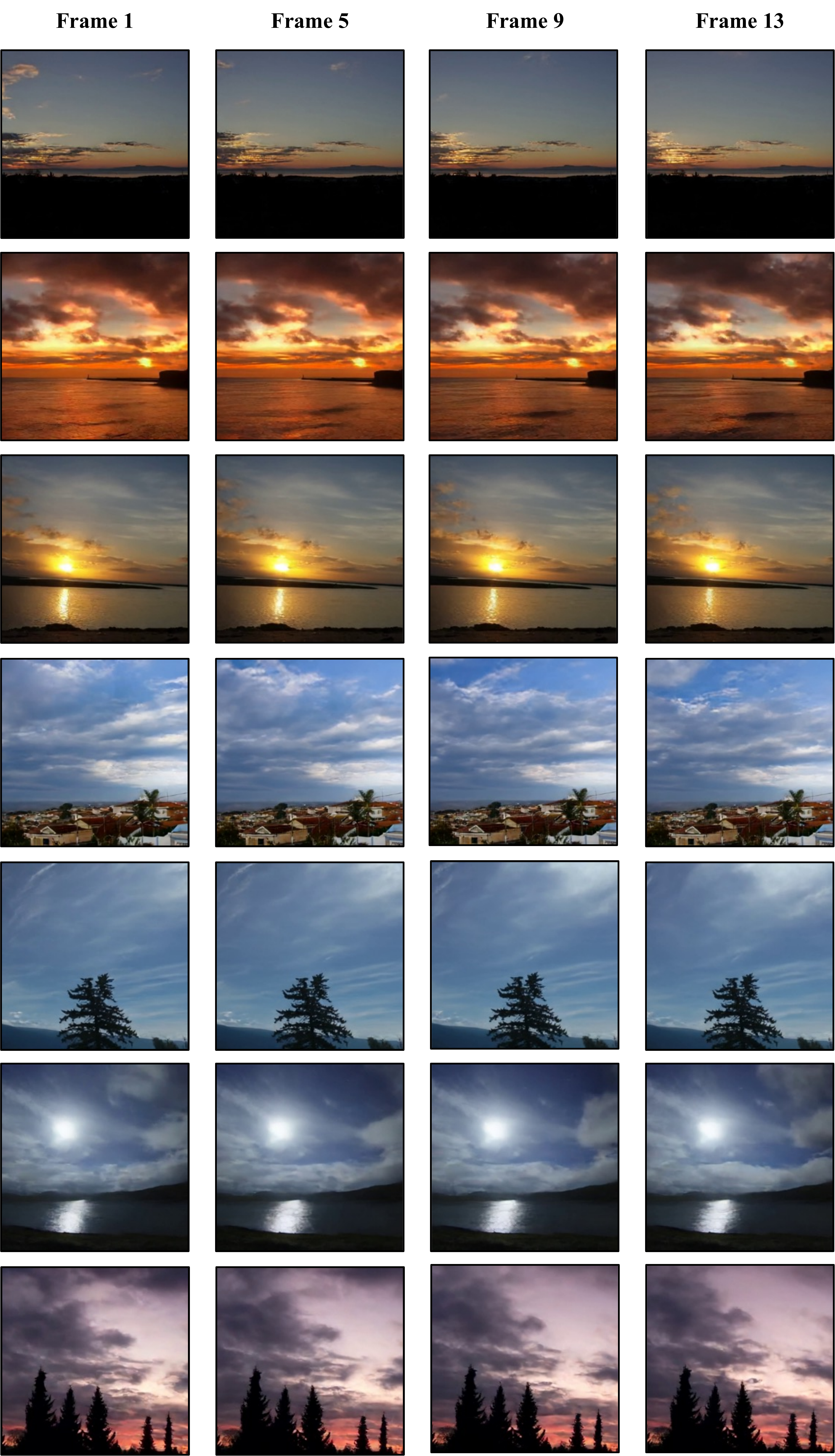}
\caption{More video examples generated by Latte-XL/2 based on {\ourmethod}-B.}
\label{fig:more_generated_videos}
\end{figure}

 We present more videos generated by the latent video generation model (\ie Latte-XL/2) built upon our {\ourmethod}-B in Fig.~\ref{fig:more_generated_videos}. Based on Fig.~\ref{fig:more_generated_videos}, we can see that the latent video generation trained on the latent space from our {\ourmethod} is capable of generating diverse and realistic videos, further demonstrating that our proposed conditioned diffusion-based video tokenizer ({\ourmethod}) can serve as a powerful foundation for video generation.

\subsection{Performance Sensitivity to the Latent Dimension}
\label{sec:appendix_latent_dim}

\begin{figure}[h!]
\vspace{30pt}
\centering
\includegraphics[width=0.5\linewidth,clip,trim=0 0 0 0]{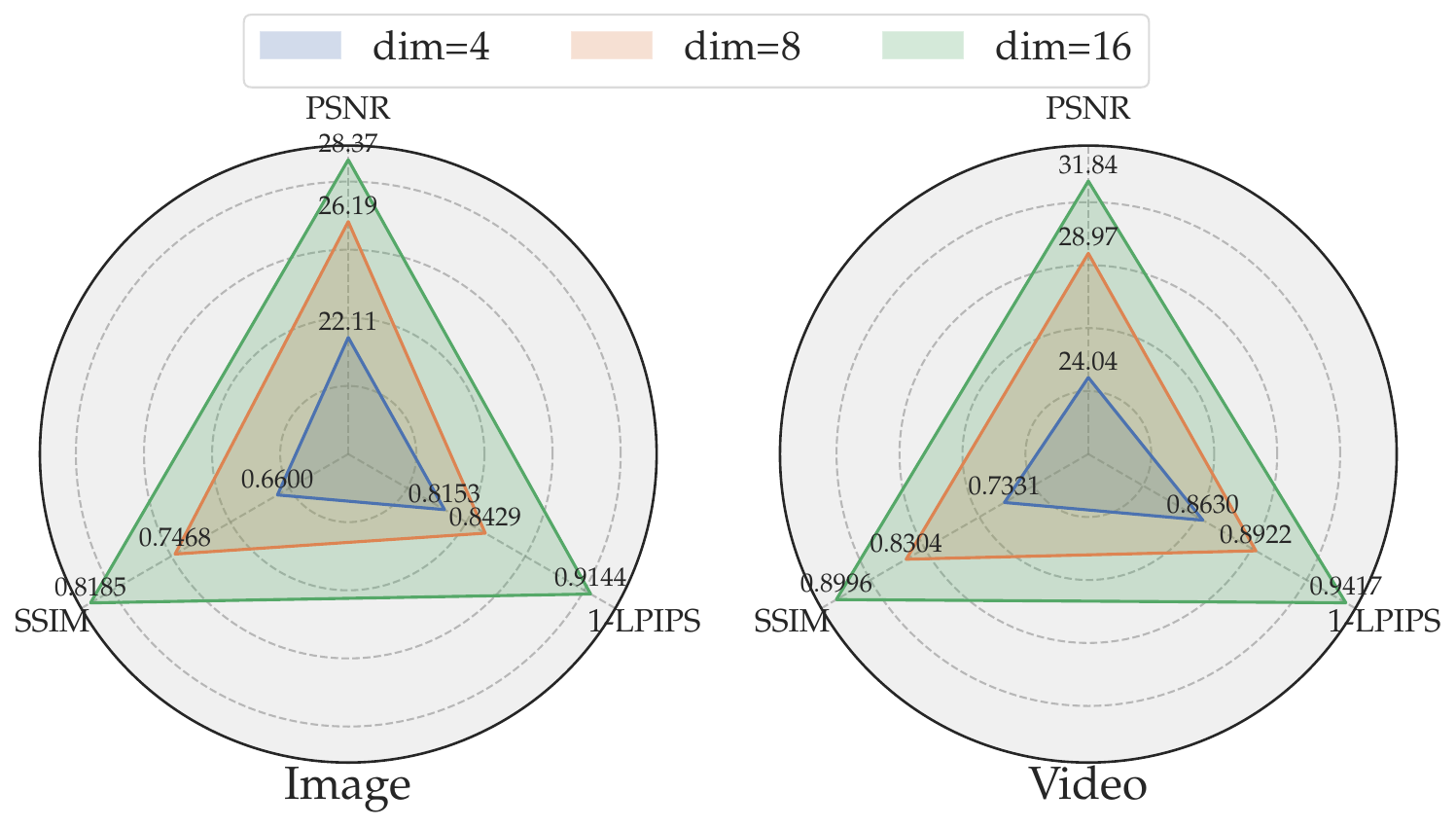}
\caption{Latent dimension impact on the reconstruction performance in terms of PSNR ($\uparrow$), SSIM ($\uparrow$), 1-LPIPS ($\uparrow$).}
\label{fig:different_dims}
\end{figure}

In this study, we examine how the dimension of latent representation affects the performance of our proposed video tokenizer. 
Similar to the experiments regarding hyper-parameter sensitivities in the main paper, we continue to evaluate models based on {\ourmethod}-S. Images are evaluated at their original resolution, while videos are evaluated at a resolution of 17~$\times$~256~$\times$~256.
By using the {\ourmethod}-S model configuration, we vary only the latent representation dimension with values of $\{4, 8, 16\}$. Each model is trained for 100k steps using four A100 GPUs. The results for image and video reconstruction are shown in Fig.~\ref{fig:different_dims}. The model with a latent dimension of 16 perform best across all metrics for both image and video data, while the model with a dimension of 8 consistently perform second best. We can observe a strong positive correlation between our method's reconstruction performance and the latent dimension. This correlation likely arises because higher latent dimension allows for more detailed encoding of the data, enabling the model to better capture and reconstruct complex patterns.

\section{Notations}
\label{sec:notations}
We summarize the notations used in this paper in Table~\ref{tab:notations} to facilitate reading.

\begin{table}[h]
    \centering
    \caption{Notations.}
    \begin{tabular}{c|l}
    \toprule
        \textbf{Notation} &
        \textbf{Description} \\
        \midrule
        $\mathcal{E}$ & the encoder of the tokenizer \\
        $\varphi$ & learnable parameters for the encoder $\mathcal{E}$ \\
        $\mathcal{D}$ & the decoder of the tokenizer \\
        $\varphi$ & learnable parameters for the decoder $\mathcal{D}$ \\
        $\mathbf{z}$ & encoded latent representation \\
        $\mathbf{V}$ & the input video \\
        $\hat{\mathbf{V}}$ & the reconstructed video  \\
        $\mathbf{V}_0$ & the input clean video in the context of diffusion framework \\
        $\hat{\mathbf{V}}_{0}$ & the predicted clean video\\
        $t$ & a specific timestep in the context of diffusion framework \\ 
        $T$ & the total timestep number in the context of diffusion framework \\
        
        $\beta_{t}$ & the noise scheduler in diffusion \\
        $\alpha_{t}$ & $\alpha_{t}=1-\beta_{t}$ \\
        $\bar{\alpha}_{t}$ & $\bar{\alpha}_{t}=\Pi_{\tau=1}^{t}\alpha_{\tau}$ \\
        $\epsilon$ & injected noise \\
        $\mu_{\theta}$ & the parameterized network for predicting the posterior mean \\
        $\epsilon_{\theta}$ & the parameterized network for predicting the injected noise \\
        $\mathcal{V}_{\theta}$ & the parameterized network for predicting the clean video \\
        $N$ & the sampling step number using DDIM in decoding phase \\
        \bottomrule
    \end{tabular}
    \label{tab:notations}
\end{table}

\clearpage

\end{document}